\documentclass[runningheads]{llncs}
\usepackage{graphicx}
\usepackage{placeins}
\usepackage{amssymb}
\usepackage{amsmath}
\usepackage{url}
\usepackage{xcolor}
\usepackage{verbatim} 
\usepackage{adjustbox} 
\usepackage{multirow} 
\usepackage{hyperref}
\usepackage{booktabs}
\usepackage{marvosym} 

\begin{document}
\title{SpaRG: Sparsely Reconstructed Graphs for Generalizable fMRI Analysis}
\author{
Camila Gonz\'alez\inst{*}\orcidID{0000-0002-4510-7309}\textsuperscript{\Letter} \and
Yanis Miraoui\inst{*} \and
Yiran Fan \and \\
Ehsan Adeli \and
Kilian M. Pohl}

%
\authorrunning{C. Gonz\'alez et al.}
%
\institute{Stanford University, Stanford, CA 94305, USA\\
\{camgonza,ymiraoui\}@stanford.edu\\
* These authors had an equal contribution.}

\maketitle              
\begin{abstract}
Deep learning can help uncover patterns in resting-state functional Magnetic Resonance Imaging (rs-fMRI) associated with psychiatric disorders and personal traits. Yet the problem of interpreting deep learning findings is rarely more evident than in fMRI analyses, as the data is sensitive to scanning effects and inherently difficult to visualize. We propose a simple approach to mitigate these challenges grounded on sparsification and self-supervision. Instead of extracting post-hoc feature attributions to uncover functional connections that are important to the target task, we identify a small subset of highly informative connections during training and occlude the rest. To this end, we jointly train a (1) sparse input mask, (2) variational autoencoder (VAE), and (3) downstream classifier in an end-to-end fashion. While we need a portion of labeled samples to train the classifier, we optimize the sparse mask and VAE with unlabeled data from additional acquisition sites, retaining only the input features that generalize well. We evaluate our method -- \textbf{Spa}rsely \textbf{R}econstructed \textbf{G}raphs (\textbf{SpaRG}) -- on the public ABIDE dataset for the task of sex classification, training with labeled cases from 18 sites and adapting the model to two additional out-of-distribution sites with a portion of unlabeled samples. For a relatively coarse parcellation (64 regions), SpaRG utilizes only 1\% of the original connections while improving the classification accuracy across domains. Our code can be found at \url{github.com/yanismiraoui/SpaRG}. 

\end{abstract}
\section{Introduction}

Resting-state functional Magnetic Resonance Imaging (rs-fMRI) has yielded valuable insights into the neural underpinnings of psychiatric disorders and individual traits, facilitating a deeper understanding of shared brain activity patterns among affected individuals \cite{zhao2021adolescent}. Yet fMRIs, which comprise hundreds of volumes per scan at a low spatial resolution, are difficult for humans to interpret. The preferred way to analyze functional connectomes is via two-dimensional matrices depicting the correlation of Blood Oxygen Level Dependent (BOLD) signals between brain regions during the scanning period \cite{buxton2009introduction}. While this significantly eases interpretation, it still requires reading the connections between dozens to hundreds of brain regions. Selecting an appropriate parcellation granularity that is sufficiently precise to capture the relevant signal yet simple enough to uncover neural underpinnings and prevent model overfitting is hence critical \cite{dadi2020fine,schaefer2018local}.

Deep learning models have achieved state-of-the-art results in detecting personal characteristics from rs-fMRIs at the subject level \cite{cui2022braingb,kan2022fbnetgen,li2019graph,li2021braingnn}. Coupled with interpretability methods, such as ROI-selection pooling layers \cite{li2021braingnn}, these models can uncover brain regions and connections that are highly indicative of the target. Graph Attention Networks (GATs) have also emerged as a strategy to identify informative features by leveraging the self-attention mechanism of transformers \cite{nerrise2023explainable,yin2021graph}. However, feature attribution and attention values are continuous and can vary widely between predictions. While these strategies provide individual-level model explanations, they do not reduce the number of functional connections considered by the model and are, therefore, often difficult to interpret. Identifying connections that generalize to unseen domains is even more challenging \cite{lee2021meta,wang2020identifying}. In this work, we take a different approach from calculating attributions post-hoc by \emph{learning a small set of generalizable neural connections and guaranteeing that all predictions emerge solely from this small feature set}.

We propose \textbf{Sparsely Reconstructed Graphs (SpaRG)}, an end-to-end method that jointly trains a sparse input mask, a self-supervised variational autoencoder, and a classifier (Fig. \ref{fig:methodology}). During training, we sparsify the rs-fMRI correlation matrices by multiplying them with a mask $\mathbf{\mathcal{M}}$. The sparse input $\mathbf{x}' = \mathcal{M} \odot \mathbf{x}$ is reconstructed by a variational autoencoder (VAE). The reconstructed functional connectomes are then the input of a Graph Convolutional Network (GCN), which predicts the outcome. As the sparsification and VAE objectives require no ground truth labels, they can be optimized with data from unlabeled sites. This encourages the sparse mask to occlude connections that are susceptible to the acquisition shift, as these comprise a large reconstruction error. Meanwhile, the supervised classification loss training the GCN preserves connections that are informative to the classification objective.

We evaluate our method on the task of sex classification from rs-fMRIs for the public \emph{ABIDE} \cite{di2014autism} dataset and explore two levels of atlas granularity, namely the 64- and 1024-dimensional \emph{Dictionaries of Functional Modes (DiFuMo)} \cite{dadi2020fine}, which were trained on millions of fMRI volumes acquired over 27 studies. Our empirical results confirm that learning a mask and unsupervised model jointly results in a set of functional connections that are informative for downstream classification and robust across acquisition sites. In fact, SpaRG can retain and \emph{even improve} classification accuracy despite acquisition differences while occluding up to 99\% of the connectomes. The resulting feature sets are consistent across validation folds and parcellation schemes and highlight connections previously identified as relevant for sex classification in the literature.

\section{Related Work}

Previous work supports the benefits of sparsification for countering the curse of dimensionality in fMRI analyses. For example, masking the 70\% lowest correlations has resulted in improved detection accuracy of brain disorders from rs-fMRI \cite{zhang2024preserving}. Popular regularizers for reducing the feature space during training include \emph{Lasso} \cite{tibshirani1996regression}, \emph{ElasticNet} \cite{zou2005regularization}, \emph{Frobenius} \cite{krauthgamer2023comparison} and the \emph{k-support Norm} \cite{argyriou2012sparse,gkirtzou2013fmri}. Sparsification can also increase the consistency of connectivity patterns across individuals \cite{ng2012novel}. Other methods take into account the correlation between predictors \cite{ng2011generalized} or patterns that arise from different fMRI tasks \cite{rao2013sparse}. 

Similar to our approach, Ahmadi et al. \cite{ahmadi2021deep} utilize a sparse autoencoder and thresholding to identify relevant connections for Alzheimer's Disease diagnosis. Other self-supervised approaches have been used for pre-training an encoder on fMRI data \cite{malkiel2022self} and extracting subject-specific functional modes from raw fMRIs \cite{li2023computing}. For instance, Zhao et al. \cite{zhao2019variational} leverage a VAE for clustering connectivity patterns in dynamic connectome analysis and outlier detection.

We are, to our knowledge, the first to propose an end-to-end semi-supervised sparsification process operating directly on correlation matrices. Our method makes no assumptions about the data-generating process and leverages unlabeled samples, resulting in robust and interpretable downstream classifiers.

\section{Methodology}

\begin{figure}[!t]
\centering
\includegraphics[width=1\textwidth]{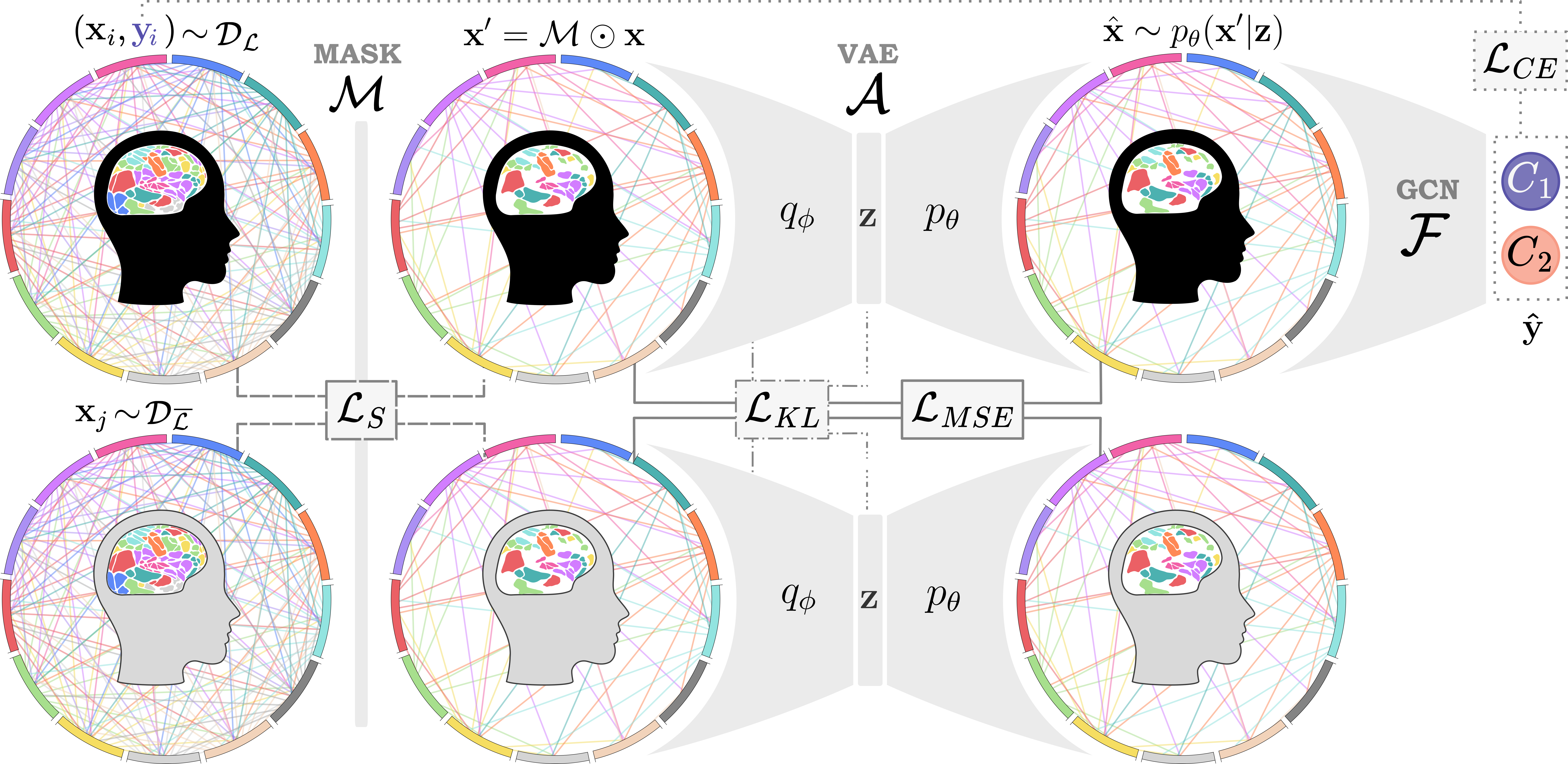}
\caption{\textbf{\emph{SpaRG:}} a sparse mask $\mathcal{M}$, variational autoencoder (VAE) $\mathcal{A}$  that reconstructs the sparse inputs, and graph convolutional network (GCN) classifier $\mathcal{F}$ are trained in an end-to-end fashion to learn a subset of robust and informative functional connections. We interleave supervised training of the GCN with self-supervised steps, where we optimize the sparsification and autoencoding losses.} 
\label{fig:methodology}
\end{figure}

In the following, we outline our learning scenario and the key components of our method, which are visualized in Fig.~\ref{fig:methodology}.

After processing the fMRIs, registering them to a common atlas, and clustering voxels into $k$ parcels, we calculate the Pearson correlation between pairwise time courses to obtain matrices of the form $\mathbf{x} \in \mathbb{R}^{k \times k}$. In our setting, labeled data is only available from a subset of sites but we have access to some unlabeled train cases from all sites -- a common scenario when performing domain adaptation. We thus have \emph{two training sets} originating from different distributions: a labeled set $\mathcal{D_L}$ with $n$ input-label pairs for our classification objective $\mathcal{D_L} = \{ (\mathbf{x}_1, \mathbf{y}_1), \ldots, (\mathbf{x}_n, \mathbf{y}_n) \}$, and a second set, smaller, set $\mathcal{D_{\ \overline{L}}}$ containing only $m$ correlation matrices $\mathcal{D_{\ \overline{L}}} = \{ \mathbf{x}_1, \ldots, \mathbf{x}_m\}$. 

\textbf{Our goal is two-fold:} we wish to \textbf{(a)} make accurate predictions $\hat{\mathbf{y}}$, generalizing well across acquisition conditions and \textbf{(b)} learn a sparse mask $\mathcal{M}$ that highlights a subset of features highly relevant for our task. Our process optimizes three objectives: \emph{sparsification, reconstruction, and classification}.

\subsection{Sparsification: $\textbf{x} \rightarrow \mathbf{x}'$}

Central to our approach is the \textbf{trainable sparse mask $\mathcal{M}$}. During the learning process, $\mathcal{M} \in \mathbb{R}^{k \times k}$ has real-valued entries $m_{i, j} = \left [ 0, 1 \right ]$. After training, we binarize $\mathcal{M}$ based on whether $m_{i, j}>\theta$ for a threshold $\theta$ based on the percentage of matrix entries to occlude. For encouraging sparsity in $\mathcal{M}$ during training we utilize \emph{ElasticNet}~(Eq. \ref{eq:elasticnet}), which combines Lasso and Ridge penalties.

\begin{equation}
\mathcal{L}_S = \lambda \sum_{i,j} |m_{i,j}| + \frac{1-\lambda}{2} \sum_{i,j} m_{i,j}^2
\label{eq:elasticnet}
\end{equation}

Applying the Hadamard product between each input $\mathbf{x}$ and $\mathcal{M}$ results in a \emph{sparsified correlation matrix} $\mathbf{x}' = \mathbf{x} \odot \mathcal{M}$. Note that this operation occurs \emph{in the first step of the forward pass} (see Fig.~\ref{fig:methodology}).

\subsection{Reconstruction: $\mathbf{x}' \rightarrow \hat{\mathbf{x}}$}

Utilizing inputs $\mathbf{x}$ from both $\mathcal{D_{L}}$ and $\mathcal{D_{\ \overline{L}}}$, we learn to \emph{reconstruct the sparse correlation matrix $\mathbf{x}'$ into $\hat{\mathbf{x}}$} with a variational autoencoder $\mathcal{A}$. Our objective here is to minimize the reconstruction $\mathcal{L}_{MSE}$, as well as the Kullback-Leibler (KL) divergence that encourages the prior distribution of the latent space $\mathbf{z}$ to follow a standard normal distribution $\mathbf{z} \sim \mathcal{N}(0, 1)$.

\begin{equation}
\mathcal{L}_{MSE} = \frac{1}{n+m} \sum_{i=1}^{n+m} \| \mathbf{x'}_i - \hat{\mathbf{x}}_i \|^{2}_{2};\ \ \mathcal{L}_{KL} = \text{KL}\left[ q_{\phi}(\mathbf{z}|\mathbf{x}) \,||\, \mathcal{N}(\mathbf{0}, \mathbf{I}) \right]
\end{equation}

This step pursues two objectives. First, by learning a structured latent space with cases from labeled and unlabeled sites, we encourage the autoencoder to learn the same posterior distribution $q(\textbf{z}|\textbf{x})$ and likelihood $p(\textbf{x}|\textbf{z})$ to reconstruct data from all domains. Second, we teach the sparse mask $\mathcal{M}$ to \emph{occlude functional connections that comprise significant differences between domains} and are therefore reconstructed incorrectly for the OOD data. Note that, as we are reconstructing the sparse input, features masked by $\mathcal{M}$ do not contribute to the reconstruction loss. Therefore, entries $\mathbf{x}_{i, j}$ that diverge significantly across sites will comprise a high reconstruction error and be subsequently occluded.

\subsection{Classification: $\hat{\mathbf{x}} \rightarrow \hat{\mathbf{y}}$}

Finally, we construct a graph from the reconstructed input $\hat{\mathbf{x}}$ and train a Graph Convolutional Network (GCN) with cross-entropy loss $\mathcal{L}_{CE}$.

We have described this process sequentially following the steps of a forward pass. However, we minimize all loss terms jointly in an end-to-end manner (Eq.~\ref{eq:total_loss}). Specifically, we perform one training step with $\mathcal{D_L}$ and one with $\mathcal{D_{\ \overline{L}}}$. In the second case, we set the classification loss $\mathcal{L}_{CE}$ to zero.

\begin{equation}
    \mathcal{L} = \lambda_1 \mathcal{L}_{S} + \lambda_2 \mathcal{L}_{MSE} + \lambda_3 \mathcal{L}_{KL} + \lambda_4 \mathcal{L}_{CE}
\label{eq:total_loss}
\end{equation}

By minimizing the joint loss, we learn a sparsification $\mathcal{M}$ that only preserves a fraction of functional connections $\mathbf{x}_{i, j}$ that are informative for our objective.

\section{Experimental Setup}

\subsection{Dataset and data preparation} 

We evaluate SpaRG on the public \emph{Autism Brain Imaging Data Exchange (ABIDE)} \cite{di2014autism} dataset, which provides a rich basis for comparison with established baselines. The data comprises rs-fMRIs from individuals with autism spectrum disorder and healthy controls acquired at 20 sites. Our in-distribution (ID) data (F: 50, M: 386; 17.87 $\pm$ 8.29) consists of controls without autism spectrum disorder from 18 sites. Cases from sites KKI and NYU form our out-of-distribution (OOD) dataset (F: 50, M: 189; 14.02 $\pm$ 6.20), which differs from the ID data in terms of acquisition site, age, and sex distribution. We perform five-fold cross-validation, training on each run with 80\% of the ID train data (the rest is used for setting hyperparameters) and 20\% of the OOD data. We do not utilize the annotations for the 20\% OOD data, simulating a setting where only a few unlabeled cases are available from the target domain. We report the balanced accuracy on ID test data and the remaining 80\% cases from KKI \& NYU.

For obtaining connectivity matrices, we apply the \emph{Dictionaries of Functional Modes (DiFuMo)} \cite{dadi2020fine}, which define 64 or 1024 \emph{soft} brain regions capturing population-wise and individual dynamics. Dadi et al. \cite{dadi2020fine} specify, for each region, which network from the 17-network atlas by Yeo at al. \cite{yeo2011organization} the region belongs to. This allows us to compare the connectivity patterns identified by the models trained with different parcellations.

\subsection{Model architectures and baselines} 

SpaRG is composed of a VAE followed by a GCN. The VAE consists of an encoder with two 16-unit hidden layers and a decoder that mirrors this structure in reverse to reconstruct the input. The classifier has 2 GCN and 2 MLP layers, each comprising 2 units. We train models with \emph{Adam} and a learning rate of 3e-4 until convergence. Given their small size, all models can be trained in a CPU.

We compare SpaRG to multiple baselines and ablations. Alternative sparsification strategies include masking the lowest correlations (\emph{Mask-GCN}) \cite{zhang2024preserving}, \emph{LASSO} sparsification \cite{tibshirani1996regression}, \emph{ElasticNet} \cite{zou2005regularization} and the \emph{Frobenius} norm \cite{krauthgamer2023comparison}. We also compare our GCN-based classifier to the \emph{explainable, geometric, weighted-graph attention network (xGW-GAT)} \cite{nerrise2023explainable}. Finally, we report ablation results of using only labeled data (SpaRG $\mathcal{D_L}$), not utilizing any sparsification or masking (SpaRG $\overline{\mathcal{L}_S}$) and using a regular autoencoder instead of a VAE (SpaRG AE). We select hyperparameters for all methods via grid search with a validation set consisting of 20\% of the ID data. These comprise the weights of the sparsification, autoencoding, and classification terms $\lambda_i \in \left [ 0.1, 0.25, 0.5 \right ]$ and the mask binarization threshold $|\mathcal{M}| \in \left [ 0, 0.7, 0.8, 0.9, 0.95, 0.98, 0.99 \right ]$, which determines the ratio of lowest correlations to fully occlude after training.

\begin{table}[!t]
    \centering
    
\begin{tabular}{p{2.8cm}|p{1.8cm}p{1.2cm}p{0.8cm}|p{1.8cm} p{1.2cm}p{0.8cm}}
\multicolumn{1}{c}{} & \multicolumn{3}{c}{\emph{\textbf{DiFuMo 64x64}}}& \multicolumn{3}{c}{\emph{\textbf{DiFuMo 1024x1024}}} \\
             & ID & OOD & $|\mathcal{M}|$\ & ID & OOD & $|\mathcal{M}|$\ \\
\hline
    
    GCN & 76.17$\pm$2.2 & 71.77 & .00 & 77.24$\pm$2.7 & 81.82 & .00 \\
    FCN    & 73.94$\pm$4.2 & 61.24 & .00 & 78.34$\pm$2.7 & 80.45 & .00 \\
    xGW-GAT \cite{nerrise2023explainable}     & 46.89$\pm$8.2 & 29.19 & .00 & 40.12$\pm$3.5 & 43.06 & .00 \\
    \hline
    Mask-GCN \cite{zhang2024preserving}       & 76.14$\pm$2.6 & 71.17 & .70 & 76.83$\pm$3.4 & 72.40 & .70 \\
    LASSO \cite{tibshirani1996regression}     & \emph{82.10$\pm$8.4} & 14.83 & \textbf{.99} & \emph{83.74$\pm$2.4} & \textbf{84.69} & \textbf{.90} \\
    ElasticNet \cite{zou2005regularization}   & 76.97$\pm$2.8 & \emph{72.73} & .00 & 83.55$\pm$2.1 & 82.76  & \emph{.80} \\
    Frobenius \cite{krauthgamer2023comparison}& 74.24$\pm$5.9 & 56.94 & .00 & 82.55$\pm$5.0 & \emph{83.25} & \emph{.80} \\
    \textbf{SpaRG (ours)}                  & \textbf{82.40$\pm$4.5} & \textbf{85.17} & \textbf{.99} & \textbf{84.28$\pm$5.5} & 82.77 & \emph{.80} \\
\bottomrule
\end{tabular}
    \caption{Balanced accuracy, averaged over 5 cross-validation folds, for the task of sex classification on the ABIDE dataset using multiple sparsification strategies and two different parcellation granularities: 64x64 (left) and 1024x1024 (right).}
    \label{tab:results_abide}
\end{table}

\section{Results}

We start by exploring whether we can obtain a small, informative subset of brain connections that permit accurate downstream classification and compare SpaRG to existing strategies. We then conduct an ablation study where we empirically confirm that all components in our method are needed. Finally, we make a visual inspection of the functional connections selected by our method for both atlases.

\subsection{The role of sparsification in classification accuracy}

In Table \ref{tab:results_abide}, we compare the balanced accuracy of our base GCN model (top) with multiple classifier architectures and sparsification strategies for two atlas granularities: 64$\times$64 and 1024$\times$1024. We train only with the controls of ID sites and 20\% of the KKI and NYU data as auxiliary OOD unlabeled samples. Before delving into sparsification, we compare three deep learning architectures, namely a GCN, a GAT, and a 4-layered fully connected network (FCN). The GCN obtains the best results, so we proceed with this model as our choice of classifier.

With respect to sparsification, when we utilize the course $64\times64$ parcellation (left side of the table), most approaches improve classification accuracy on ID data. This supports previous findings on the effectiveness of sparsification to counter the curse of dimensionality in fMRI analysis \cite{zhang2024preserving}. However, this only translates to higher OOD accuracy for SpaRG, which leverages a small subset of unlabeled scans from OOD sites. In column $|\mathcal{M}|$, we report the best ratio of occluded connections for each approach, selected on ID validation data. Those connections are occluded after training. Only Mask-GCN, Lasso and SpaRG perform best when occluding a large portion of the connections. The fine-grained $1024\times1024$ parcellation strategy (right side) is less susceptible to acquisition changes, as reflected in the higher accuracy on OOD data for all methods. This is potentially due to the fine-grained functional modes being more noisy and distinct between individuals \cite{dadi2020fine}, preventing the network from overfitting to scanning peculiarities during the training process. In general, utilizing the higher-dimensional matrices coupled with sparsification and post-training occlusion obtains the most reliable results.

\begin{figure}[!t]
\centering
\includegraphics[width=1\textwidth]{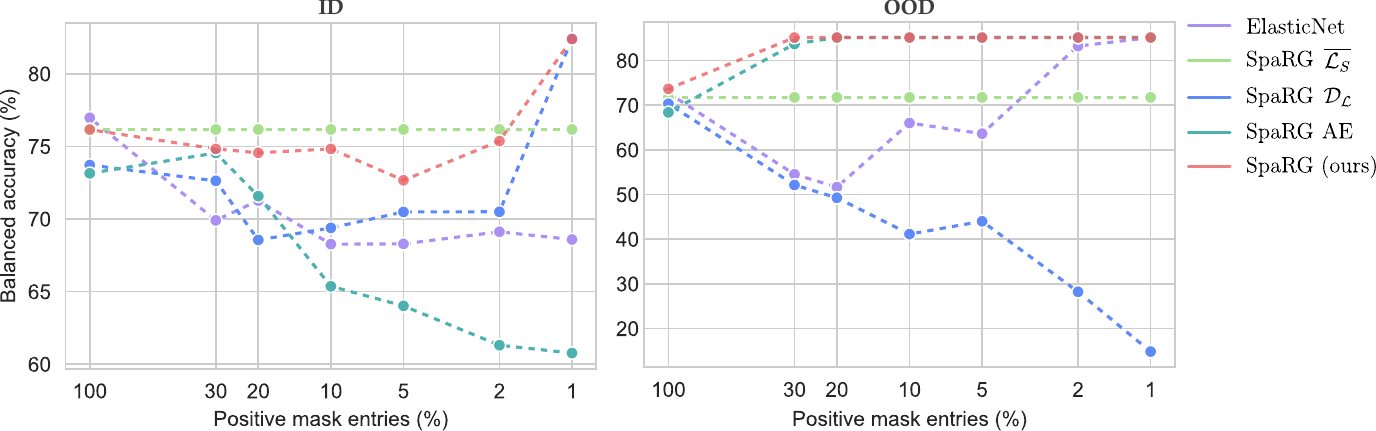}
\caption{Balanced accuracy on the ID and OOD sites for different levels of occlusion.} 
\label{fig:thresholds_all}
\end{figure}

\subsection{Self-supervision promotes generalizable occlusion}

Table \ref{tab:results_ablation} summarizes our ablation study of SpaRG. First, we explore a variant that does not perform any sparsification or masking (SpaRG $\overline{\mathcal{L}_S}$). In this setting, the VAE alone does not alleviate the effect of the distribution shift, as shown in the low accuracies for OOD data. We further demonstrate that using unlabeled data improves generalization as opposed to only leveraging labeled ID cases (SpaRG $\mathcal{D_L}$). Finally, we establish that a VAE -- which shapes the latent space to follow a standard normal -- is preferable over a regular autoencoder (SpaRG AE). 

Beyond finding a solution for a specific occlusion threshold, we conduct an analysis of multiple specification options for the 64x64 atlas. Fig. \ref{fig:thresholds_all} corroborates that SpaRG, grounded in self-supervised reconstruction, helps guide the sparsification for multiple thresholds.

\begin{table}[!t]
    \centering
    
\begin{tabular}{p{2.8cm}|p{1.8cm}p{1.2cm}p{0.8cm}|p{1.8cm} p{1.2cm}p{0.8cm}}
\multicolumn{1}{c}{} & \multicolumn{3}{c}{\emph{\textbf{DiFuMo 64x64}}}& \multicolumn{3}{c}{\emph{\textbf{DiFuMo 1024x1024}}} \\
             & ID & OOD & $|\mathcal{M}|$\ & ID & OOD & $|\mathcal{M}|$\ \\
\hline
    SpaRG $\overline{\mathcal{L}_S}$        & 76.17$\pm$4.4 & 71.77  & .00 & 83.20$\pm$1.4 & 29.19 &	\textbf{.99} \\
    SpaRG $\mathcal{D_L}$                  & \textbf{82.40$\pm$4.5} & 14.83 &  \textbf{.99} & \emph{84.02$\pm$2.4} & \emph{85.16} &	\textbf{.99} \\
    SpaRG AE              & 74.55$\pm$4.1 & \emph{83.73} & .70 & 84.01$\pm$4.3 & \textbf{85.17} &		.98 \\
    \textbf{SpaRG (ours)}                  & \textbf{82.40$\pm$4.5} & \textbf{85.17} & \textbf{.99} & \textbf{84.28$\pm$5.5} & 82.77 & .80 \\
    \bottomrule
\end{tabular}

    \caption{Ablative testing of the different components making up SpaRG.}
    \label{tab:results_ablation}
\end{table}

\subsection{Qualitative examination of the preserved functional connections}

\begin{figure}[!h]
\centering
\includegraphics[width=1\textwidth]{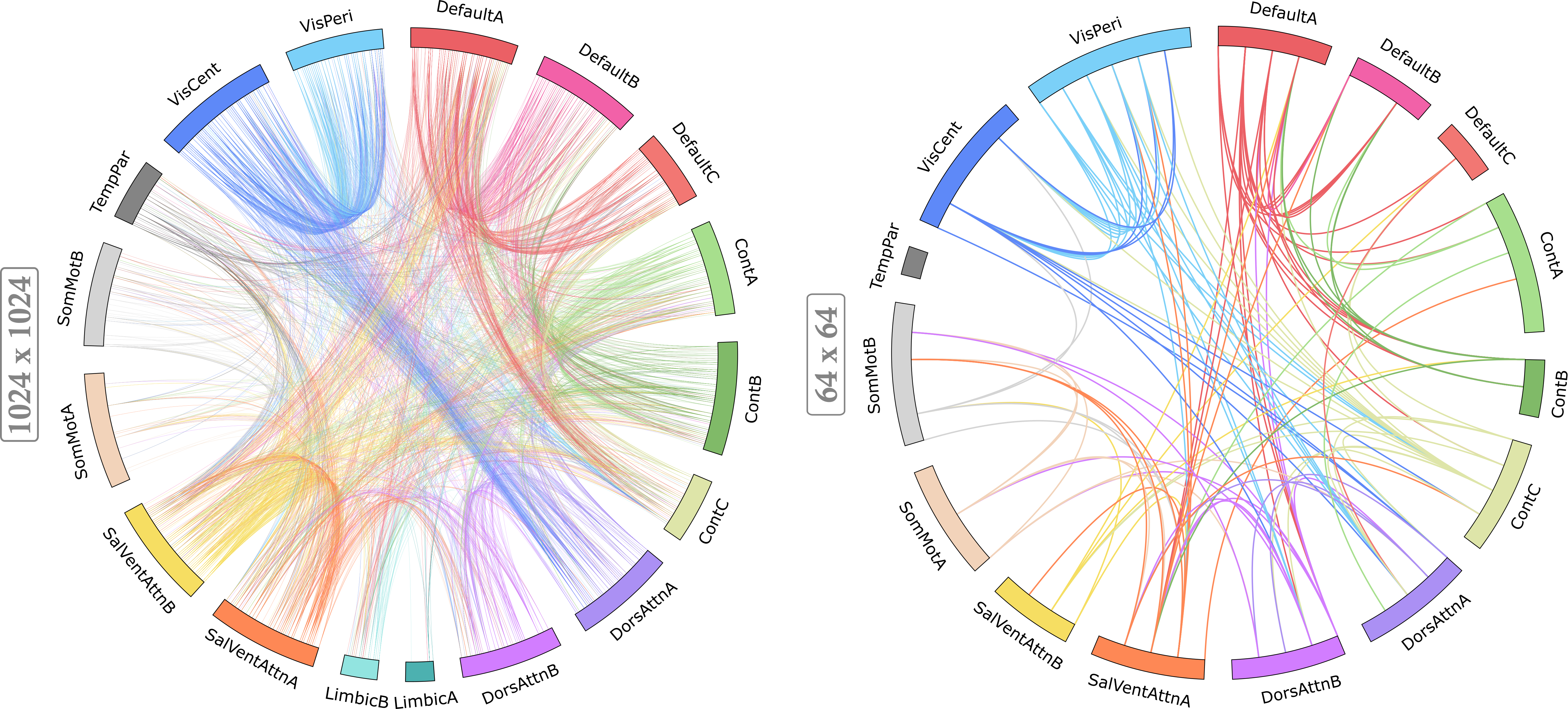}
\caption{Functional connections preserved by \emph{SpaRG} for two parcel granularities, mapped to the 17-network atlas \cite{yeo2011organization}. Similar connections are preserved by both models, highlighting connectivity involving the visual and default mode networks.} 
\label{fig:non_masked_connections}
\end{figure}

Fig. \ref{fig:non_masked_connections} shows the connections preserved by SpaRG for models trained with both parcellation granularities, clustered for comparison purposes into the networks of the Yeo et al. \cite{yeo2011organization} atlas following Dadi et al. \cite{dadi2020fine}. A visual inspection of the connectivity between networks demonstrates that similar patterns are learned by both models. Evidently, for classifying the sex from rs-fMRI, the models utilize connections that implicate visual and attention functions and the default mode network, supporting previous findings \cite{gadgil2020spatio,muller2018influences}. In this work, we focused on the well-understood task of sex classification, which allowed us to examine the potential and limitations of SpaRG beyond domain-specific design choices. Our results indicate that self-supervised sparsification can potentially allow a better exploration of the underlying mechanisms of psychiatric disorders, as we will explore in additional settings in future work.

\section{Conclusion}

Functional MRI connectivity data holds immense potential for advancing the understanding of psychiatric and neurodegenerative disorders. Yet the intrinsic difficulty in interpreting high-dimensional correlation matrices and the small reproducibility of findings across acquisition sites and populations introduce significant hurdles. We propose an alternative avenue to observing subject-level feature attributions, namely learning a sparse mask that occludes uninformative functional connections alongside a VAE that identifies connections stable across distribution shifts through self-supervision. Optimizing these components and a downstream classifier jointly allows us to find a subset of up to 1\% the size of the original correlation matrices while preserving or improving classification accuracy. These findings highlight the potential of self-supervised sparsification for increasing the interpretability of fMRI analyses.

\subsubsection{Acknowledgement}
The work was partly funded by the U.S. National Institutes of Health (NIH) grants (DA057567), Stanford HAI Google Cloud Credit, the DGIST Joint Research Project, and the 2024 Stanford HAI Hoffman-Yee Grant.

%
%
%
\bibliographystyle{splncs04}
\bibliography{references}

\end{document}